\documentclass{article} 
\usepackage{iclr2026_conference,times}

\usepackage{amsmath,amsfonts,bm}

\def\eqref#1{equation~\ref{#1}}

\def\1{\bm{1}}

\DeclareMathAlphabet{\mathsfit}{\encodingdefault}{\sfdefault}{m}{sl}
\SetMathAlphabet{\mathsfit}{bold}{\encodingdefault}{\sfdefault}{bx}{n}

\usepackage{hyperref}
\usepackage{url}
\usepackage{graphicx}
\usepackage{subcaption}
\usepackage{float}

\title{Hierarchical Reasoning Models: Perspectives and Misconceptions
\\
\small
}

\author{Renee Ge\footnotemark[1] \,\,$^{1,4}$ , \, \textbf{Qianli Liao}\footnotemark[1]\,\, $^{1,2,3}$ ,\&\, \textbf{Tomaso Poggio}\,\,\, $^{1,3,4}$
\\
$^1$ Center for Brains, Minds, and Machines \\
$^2$ K. Lisa Yang Integrative Computational Neuroscience (ICoN) Center \\
$^3$ McGovern Institute for Brain Research \\
$^4$ Department of Brain and Cognitive Sciences \\
MIT\\
Cambridge, MA 02139, USA \\
}
\newcommand\blankfootnote[1]{%
  \let\thefootnote\relax\footnotetext{#1}%
  \let\thefootnote\svthefootnote%
}
\blankfootnote{$*$ Equal contribution}
\iclrfinalcopy 
\begin{document}

\maketitle

\begin{abstract}
Transformers have demonstrated remarkable performance in natural language processing and related domains, as they largely focus on sequential, autoregressive next-token prediction tasks. Yet, they struggle in logical reasoning, not necessarily because of a fundamental limitation of these models, but possibly due to the lack of exploration of more creative uses, such as latent space and recurrent reasoning. An emerging exploration in this direction is the Hierarchical Reasoning Model \cite{Wang+etAl2025_hrm}, which introduces a novel type of recurrent reasoning in the latent space of transformers, achieving remarkable performance on a wide range of 2D reasoning tasks. Despite the promising results, this line of models is still at an early stage and calls for in-depth investigation.  In this work, we review this class of models, examine key design choices, test alternative variants and clarify common misconceptions.

\end{abstract}

\section{Introduction}
\label{introduction}


The Transformer architecture \citep{vaswani2017attention} has become the foundation of modern large language models (LLMs), powering systems such as BERT \citep{devlin2019bert}, PaLM \citep{chowdhery2022palm} and GPT series  \citep{brown2020language,achiam2023gpt}. Scaling these models has led to impressive progress across natural language understanding and generation tasks. Nevertheless, benchmarks specifically designed to probe reasoning --- for example, GSM8K for arithmetic reasoning \citep{cobbe2021training}, BIG-Bench for diverse challenging tasks \citep{srivastava2022beyond} and MATH for competition-level mathematics \citep{hendrycks2021measuring} --- have revealed limitations in systematic multi-step reasoning, even for the largest models.

To address these gaps,  methods have been proposed to elicit reasoning through prompting or architectural modifications. Chain-of-thought prompting \citep{wei2022chain} and scratchpad supervision \citep{nye2021show} train or prompt models to generate explicit intermediate steps, while program-aided approaches such as PAL \citep{gao2023pal} offload computation to external interpreters. At the same time, theoretical studies highlight structural limits of attention-only architectures: for example, \citet{abbe2024far} demonstrate a ``locality barrier'' that constrains global reasoning unless additional memory mechanisms are introduced. These insights echo findings from cognitive science, which emphasizes the hierarchical and compositional nature of human reasoning \citep{dedhe2023origins}. Together, this body of work motivates the exploration of architectures and training paradigms that go beyond standard token-level generation, such as latent-space reasoning and recurrent refinement, which we examine in this paper, with an emphasis on the latest popular embodiment of these two ideas ---  Hierarchical Reasoning Model (HRM) \citep{Wang+etAl2025_hrm}.

Despite the significant attention gained by HRM due to its innovative designs and impressive performance on various reasoning benchmarks, many aspects remain unclear. This paper aims to investigate some key design elements of HRM, for example: 1. Is HRM's hierarchical architecture important, in particular the L and H modules? 2. How novel is HRM's ``biologically plausible'' one-step gradient training? 3. What benefit does HRM's Adaptive Computation Time (ACT) bring? Why is there a discrepancy in HRM's training and test procedures? Through analyses of these key features, we aim to clarify common misconceptions regarding their importance and offer a clearer understanding of the model.

\newpage

\section{Latent Space Reasoning and Hierarchical Reasoning Model (HRM)}
\subsection{Latent Space vs. Token Space Reasoning}
Token-space reasoning relies on generating explicit intermediate steps as natural-language tokens, as in chain-of-thought prompting \citep{wei2022chain}. While such methods improve interpretability, they often introduce inefficiency: reasoning traces may be long, redundant, and brittle to errors in individual tokens \citep{nye2021show,kojima2022large}. Moreover, many benchmarks reveal that token-based reasoning, despite improvements, remains slow and sometimes unreliable on harder problems \citep{srivastava2022beyond,cobbe2021training}.

Latent-space reasoning takes a different approach, performing reasoning operations within the model’s hidden representations rather than explicit text. Recent work on latent chain-of-thought and related methods \citep{hao2024training} shows that iterative refinement in latent space can yield more compact and flexible reasoning traces. These approaches reduce reliance on token generation, enabling richer intermediate states, faster inference, and in some cases improved accuracy \citep{hao2024training}. Latent-space reasoning thus provides a promising complement to explicit token-space methods.

\subsection{HRM vs. Transformers}
The Hierarchical Reasoning Model (HRM) \citep{Wang+etAl2025_hrm} is the latest popular model that performs latent-space reasoning. It introduces a dual-loop recurrent architecture inspired by cognitive science. It alternates between a fast low-level (L) module that executes candidate reasoning steps and a slower high-level (H) module that updates a more abstract plan. HRM iterates these cycles until a learned halting signal indicates completion, yielding effective depth far beyond its nominal layer count. Notably, HRM achieves strong performance on reasoning-heavy benchmarks such as Sudoku and ARC-AGI-1 \citep{chollet2019measure} without pretraining or explicit chain-of-thought supervision \citep{Wang+etAl2025_hrm}. Despite the promising ideas and concepts introduced by the paper, it is unclear how much benefit these L and H modules bring, compared to a plain transformer. In our experiments in Figure \ref{fig:sudoku_8_comparison}, we show that even without H module, a plain 8-layer L-Module only HRM performs similarly to the original 4-layer L-module, 4-layer H-module HRM. This is consistent to what is reported by \citet{arc_HRM}, which says that the plain transformer performs similarly to HRM.

\section{Recurrent Reasoning as a Type of Diffussion Model}
\subsection{HRM is a Diffusion Model}
HRM attracted much attention upon release, as indicated by its 10k Github stars \citep{HRM_github} received within only two months. Its popularity can be attributed to its unique design and training procedure, as well as its strong performance. A key novelty is that HRM frameworks its recurrence in a non-Backpropagation Through Time (BPTT) setting and thus distinguishes itself sharply from any Recurrent Neural Networks (RNNs), giving people from the traditional language processing and RNN background a fresh impression.

However, HRM can be an interesting example model whose novelty actually varies by perspective: while it is quite innovative from a traditional RNN standpoint, it appears more routine from the alternative perspective we propose below:

HRM’s training algorithm — which replaces BPTT with a ``one-step'' gradient rule — turns out to be very similar to how diffusion models are trained. In a denoising diffusion process, one repeatedly applies a neural network to gradually denoise a sample, learning to map a noisy input back to the clean data \citep{ho2020denoising,rombach2022high,Song2023Consistency}. Similarly, HRM repeatedly refines an internal latent state towards the solution. 

More formally, diffusion training uses example pairs $(x_{t1}, x_{t2})$ to teach a network $f_\theta(x_t)$ to recover the original data $x_0$ from a noisy version $x_t$, where $t1-t2$ can be small (classical diffusion model \citep{ho2020denoising,rombach2022high}) or large (consistency model \citep{Song2023Consistency}). HRM’s one-step training can be viewed the same way: the model takes an \emph{intermediate reasoning state} (analogous to $x_t$) and learns to map it directly to the final answer (analogous to $x_0$). The only substantive difference is the data: diffusion models add artificial noise to construct $x_t$, whereas HRM’s latent state is an intermediate reasoning state generated by the model. Since HRM operates on latent space and learns directly the mapping from $x_t$ to $x_0$, it corresponds precisely to ``Latent Consistency Model'' (LCM) \citep{Song2024Latent}.  

Consistency models \citep{Song2023Consistency,Song2024Latent} can generate high-quality, visually appealing images with as few as 2-4 inference steps. As we will demonstrate later, a well-trained HRM on the Sudoku task also significantly reduces its inference steps, down to as few as 2 to 4 steps, which is surprising yet somewhat expected when viewing HRM as a LCM. From a task standpoint, few-step image generation appears plausible, as humans can all perform instant visual imagination. However, few-step reasoning on Sudoku is unexpected. It is intriguing to see how these models learn to solve these seemingly unrelated tasks in a similar manner when trained using the same framework. 


\subsection{BPTT vs. Diffusion: a Biological Perspective}
A key motivation for HRM is biological plausibility: by avoiding BPTT, the authors claim a more brain-like learning rule. Once we recognize the parallel with diffusion models, it follows that any diffusion-like recurrent process could share this advantage. Diffusion models do not require backpropagation through multiple time steps; they train a feedforward mapping from one state to another independently. In HRM’s words, the one-step gradient has ``constant memory'' versus BPTT’s growing footprint. Diffusion training is inherently of this form: the model learns locally at each step without unfolding the entire recurrence in memory.  

Therefore, diffusion models offer an equally biologically plausible paradigm for deep sequential computation. Generating a sample with a diffusion model is effectively a recurrent process (iteratively refining a latent) but it is trained with a feedforward loss. This suggests that mechanisms inspired by diffusion could provide a basis for ``deep reasoning'' in the brain without requiring implausible credit assignment. In summary, HRM’s replacement of BPTT is not unique: any latent consistency or diffusion model likewise avoids BPTT, pointing to a broader potential for biologically plausible recurrent learning. Further work should formalize this connection and explore how diffusion-like updates might operate in neural circuits. 

However, it is important to note that avoiding BPTT does not equate to being fully biologically implementable. A deep feedforward model still requires credit assignment from later layers to earlier ones, where a few algorithms may be possible (e.g., \cite{liao2024self,nokland2016direct,lillicrap2016random,liao2016important}). The biological plausibility of transformer blocks, on the other hand, remains an open question.

\section{Adaptive Computation Time (ACT) in HRM}
Adaptive Computation Time (ACT), in a general sense, is a concept of allowing a model to perform different amount of computation (e.g., number of iterations) on different input data. It was proposed in \citep{graves2016adaptive} on Recurrent Neural Networks (RNN). For a feedforward network, applying some form of ACT essentially converts it into an RNN and allows it to scale its computation beyond a feedforward model and potentially achieve Turing Completeness \citep{liao2016bridging}. This is one of the core motivations of HRM, which can be interpreted as roughly equivalent to applying ACT to transformers.

Although ACT is a key design in HRM, it is not sufficiently discussed in the original paper \citep{Wang+etAl2025_hrm}.
There is a critical mismatch in the original HRM implementation: while its ACT module learns to decide when to halt during training, evaluation ignores this policy and always executes the maximum number of reasoning steps. To further explore this, we modify HRM to use its learned halting policy during evaluation as well.

There are a few different ways of performing ACT during inference: 1. baseline: apply a fixed but different number of reasoning steps from training for all examples 2. use a decision rule to decide the number of reasoning steps for every sample. We will describe them in detail in the experiments section.

\section{Experiments}
\subsection{Methods}
\label{methods}
\subsubsection{Removal of H-Module}
The original HRM alternates between two reasoning modules: a fast-updating low-level L-module and a slower high-level H module that updates once every $T$ steps. These modules maintain internal latent states $(z_L, z_H$ and are updated as:

\begin{align*}
z_L^i &= f_L(z_L^{i-1}, z_H^{i-1}, \tilde{x}; \theta_L), \\
z_H^i &= 
\begin{cases}
f_H(z_H^{i-1}, z_L^{i-1}; \theta_H) & \text{if } i \equiv 0 \pmod{T}, \\
z_H^{i-1} & \text{otherwise}.
\end{cases}
\end{align*}

In \citet{Wang+etAl2025_hrm}, the intuition of these hierarchical reasoning modules is that the H-module updates less frequently, as it is meant to process global abstractions, while the L-module updates more frequently, as it is intended to handle local reasoning. However, in our experiments, we create an expanded 8-layer L-module only HRM that updates at each timestep $i$, eliminating the biologically-inspired hierarchical design choice of maintaining two separate latent states $z_L$ and $z_H$. As shown in Figure \ref{fig:sudoku_8_comparison}, our approach achieves similar performance as the original HRM, casting doubt on how much benefit the hierarchical L and H modules bring to the HRM.

\subsubsection{ACT for both Training and Inference}

We also investigate the HRM's ACT module. The model's ACT module uses Q-learning to balance exploration (continuing reasoning) against exploitation (terminating with the current solution). At each step $m$, a Q-head outputs values for \emph{halt} and \emph{continue} as ${Q}^m = \bigl(\hat{Q}^m_{\text{halt}}, \hat{Q}^m_{\text{continue}}\bigr)$.
Halting occurs if (i) the maximum horizon $M_{\max}$ is reached or (ii)
$\hat{Q}^m_{\text{halt}} > \hat{Q}^m_{\text{continue}}$ once a stochastic minimum threshold $M_{\min}$ has been satisfied. 

Choosing \emph{halt} gives reward
$\hat{G}^m_{\text{halt}} = \mathbf{1}\{\hat{y}_m = y\}$,
while \emph{continue} yields
\[
\hat{G}^m_{\text{continue}} =
\begin{cases}
\hat{Q}^{m+1}_{\text{halt}}, & m \geq M_{\max}, \\
\max\!\bigl(\hat{Q}^{m+1}_{\text{halt}}, \hat{Q}^{m+1}_{\text{continue}}\bigr), & \text{otherwise}.
\end{cases}
\]

The loss at step $m$ combines prediction accuracy and stopping decisions:
\[
\mathcal{L}^m_{\text{ACT}} = \text{Loss}(\hat{y}_m, y) \;+\; \text{BCE}(\hat{Q}^m, \hat{G}^m).
\]

In the released HRM source code \citep{HRM_github}, however, while the ACT module is used during training, evaluation almost always executes for the full $M_{max}$ steps. This approach ignores the halting mechanism, ensuring uniform batch shapes but preventing evaluation from using the model’s learned stopping policy to its full potential. As a result, evaluation performance diverges from training performance, as the model under evaluation may take unnecessary extra steps even when a puzzle is already solved. By neglecting to use the ACT module to halt when the puzzle is already solved, potential computational efficiency gains and performance are compromised.

We modify HRM to apply halting learned during training to evaluation as well, determined by the halting logits ($\hat{Q}_{halt}$ and $\hat{Q}_{continue}$). We experiment (as shown in Figure \ref{fig:per_sample}) with halting the model in two cases: (1) when $sigmoid(\hat{Q}_{halt}) > threshold$, and (2) when $sigmoid(\hat{Q}_{halt} - \hat{Q}_{continue}) > threshold$, observing the effects of varying the halting threshold on model performance.

Evaluation proceeds until all sequences in a batch have completed, whether through the ACT module halting the sequence or by evaluating the sequence up to the max step. Rather than running all the sequences up to the maximum step $M_{max}$, as in the original HRM implementation, this modification allows evaluation to measure both task accuracy and the model’s ability to adaptively allocate computation.

\subsection{Results}

\begin{figure}[H]
    \centering
    \includegraphics[width=0.75\linewidth]{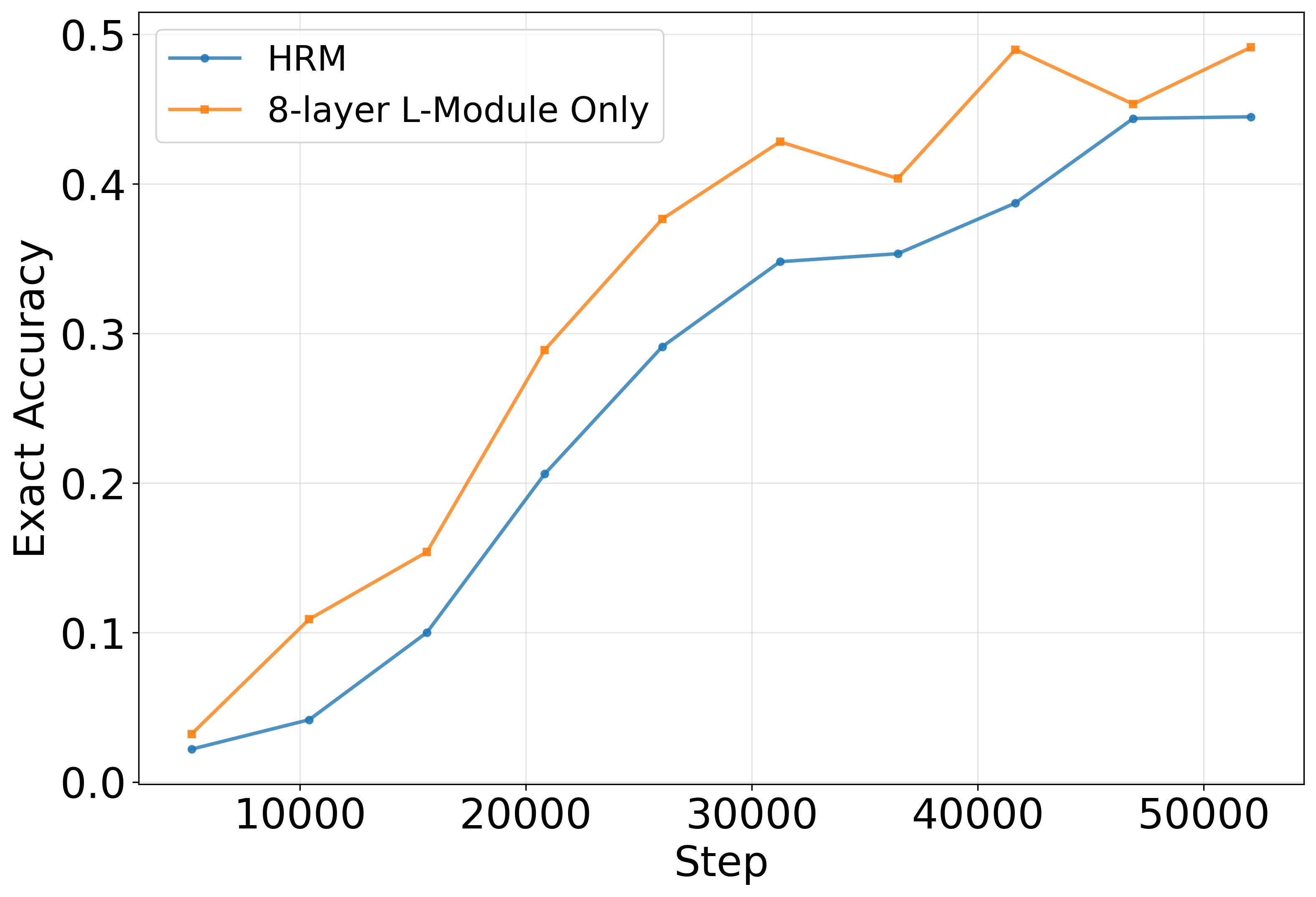}
    \caption{Comparison between original HRM (4-layer H-module and 4-layer L-module and HRM with a 8-layer L-module only (L-cycle=1). The latter is equivalent to a 8-layer plain transformer. The x axis is the number of training steps/iterations (i.e., number of minibatches). The plain transformer/L-module-only setting performs similarly or slightly better than the original HRM while at the same time runs much faster (runtime 1h 48m vs. original HRM's 4h 21m on a A100 GPU).}
    \label{fig:sudoku_8_comparison}
\end{figure}

In Figure \ref{fig:sudoku_8_comparison}, we remove the H-module in the HRM and instead expand the L-module to 8 layers.

\begin{figure}[H]
    \centering
    \includegraphics[width=\linewidth]{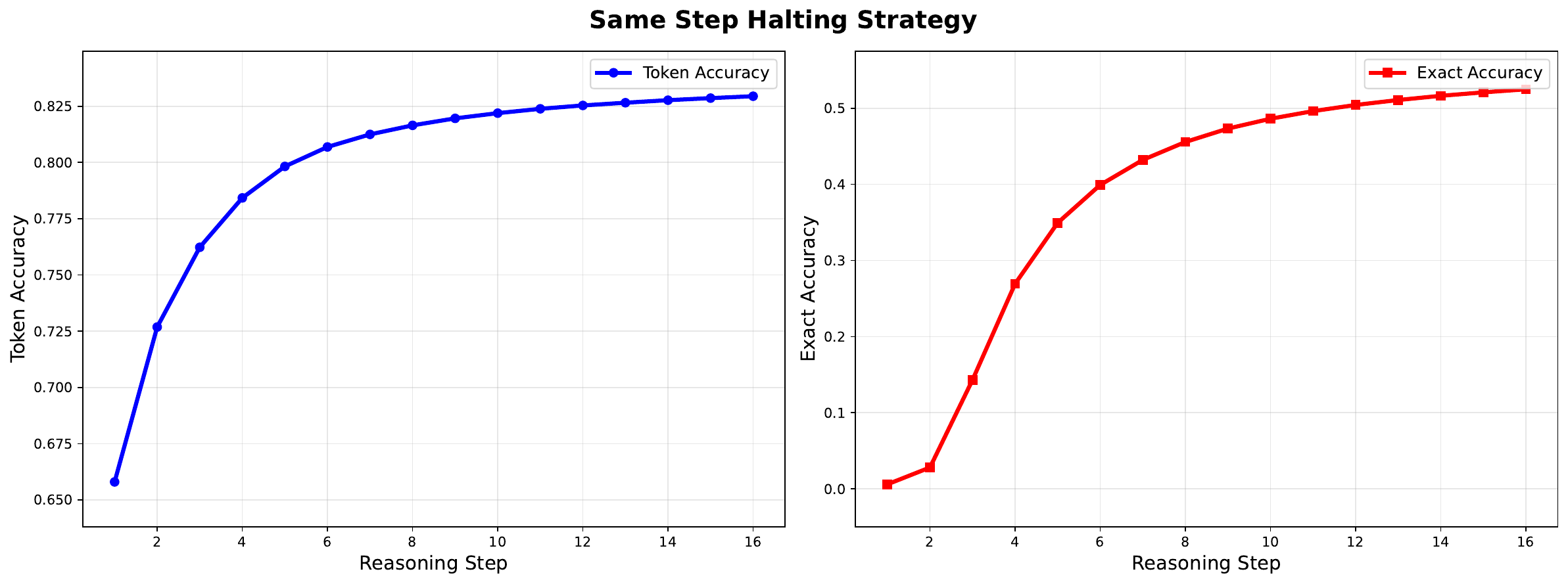}
    \caption{Reasoning with \textbf{same inference steps for all examples} on the Sudoku task. The x-axis represents the number of reasoning steps, which is kept the same across all examples. We then plot the performance curves for token accuracy and exact accuracy, with exact accuracy being 1 for a fully correct sequence and 0 otherwise for each example. The original HRM is used as the model architecture. The observation that using the same number of reasoning steps on all examples can help raise a question about the recurrent nature of the HRM paradigm. Is it truly recurrent or does it function effectively as a very deep feedforward model?
 }
    \label{fig:same_step}
\end{figure}
\begin{figure}[H]
    \centering
    \includegraphics[width=\linewidth]{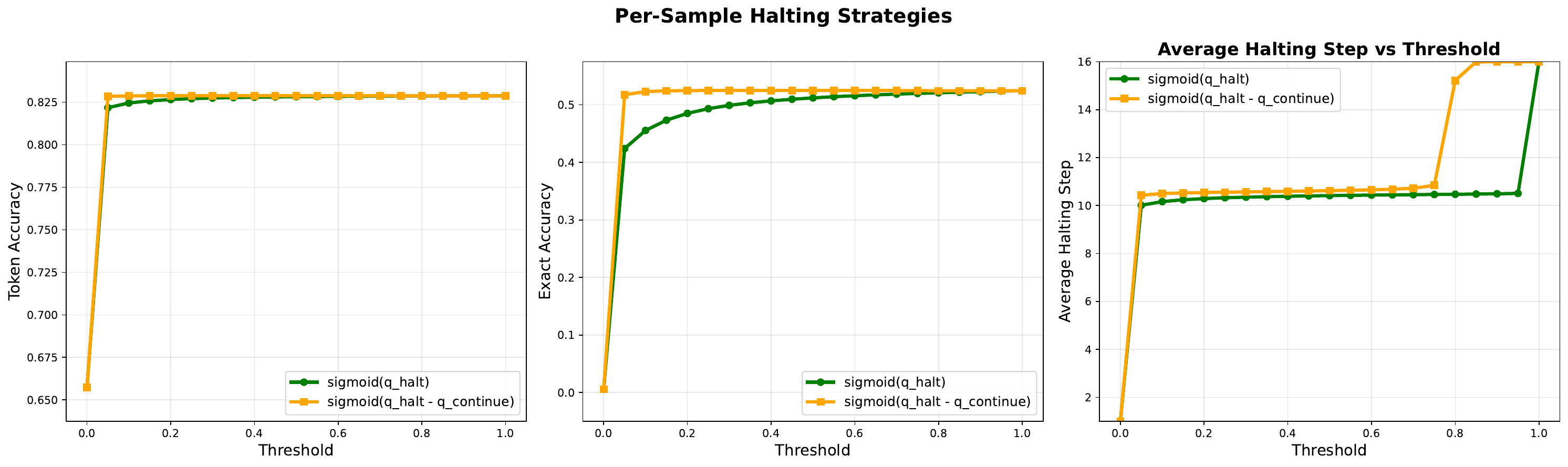}
    \caption{Sample-specific reasoning in inference with different strategies. In contrast to Figure \ref{fig:same_step}, in this setting every example has a different number of reasoning steps, determined by the halting logits ($\hat{Q}_{halt}$ and $\hat{Q}_{continue}$) generated by the model. A model decides to halt using two strategies: 1. when $sigmoid(\hat{Q}_{halt}) > threshold$. 2. when $sigmoid(\hat{Q}_{halt} - \hat{Q}_{continue}) > threshold$. The x-axes of the 3 subfigures represent the threshold. The y-axes represent token accuracy, exact accuracy and average halting steps, respectively. The original HRM is used as the model architecture.
}
    \label{fig:per_sample}
\end{figure}

\begin{figure}[H]
    \centering
    \includegraphics[width=\linewidth]{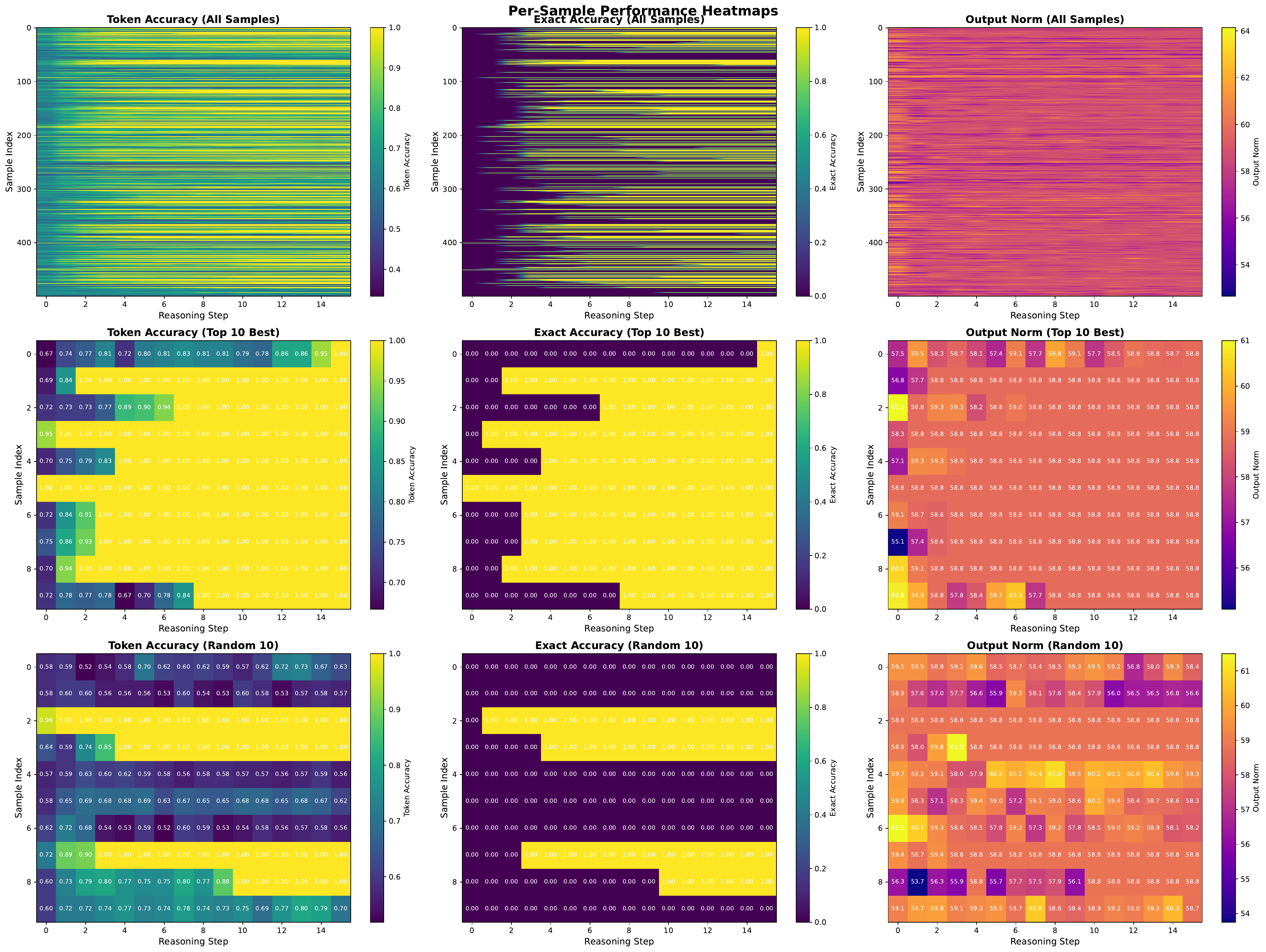}
    \caption{Example sample performances with different reasoning steps. Token accuray (left column), exact accuracy (middle column) and output norm (right column). These figures shows the performances of a selected number of examples on various reasoning steps. The second row shows the statistics from 10 examples with best performance. The third row shows the statistics from 10 random examples. }
    \label{fig:heatmaps}
\end{figure}

\begin{figure}[H]
    \centering
    \includegraphics[width=1.0\linewidth]{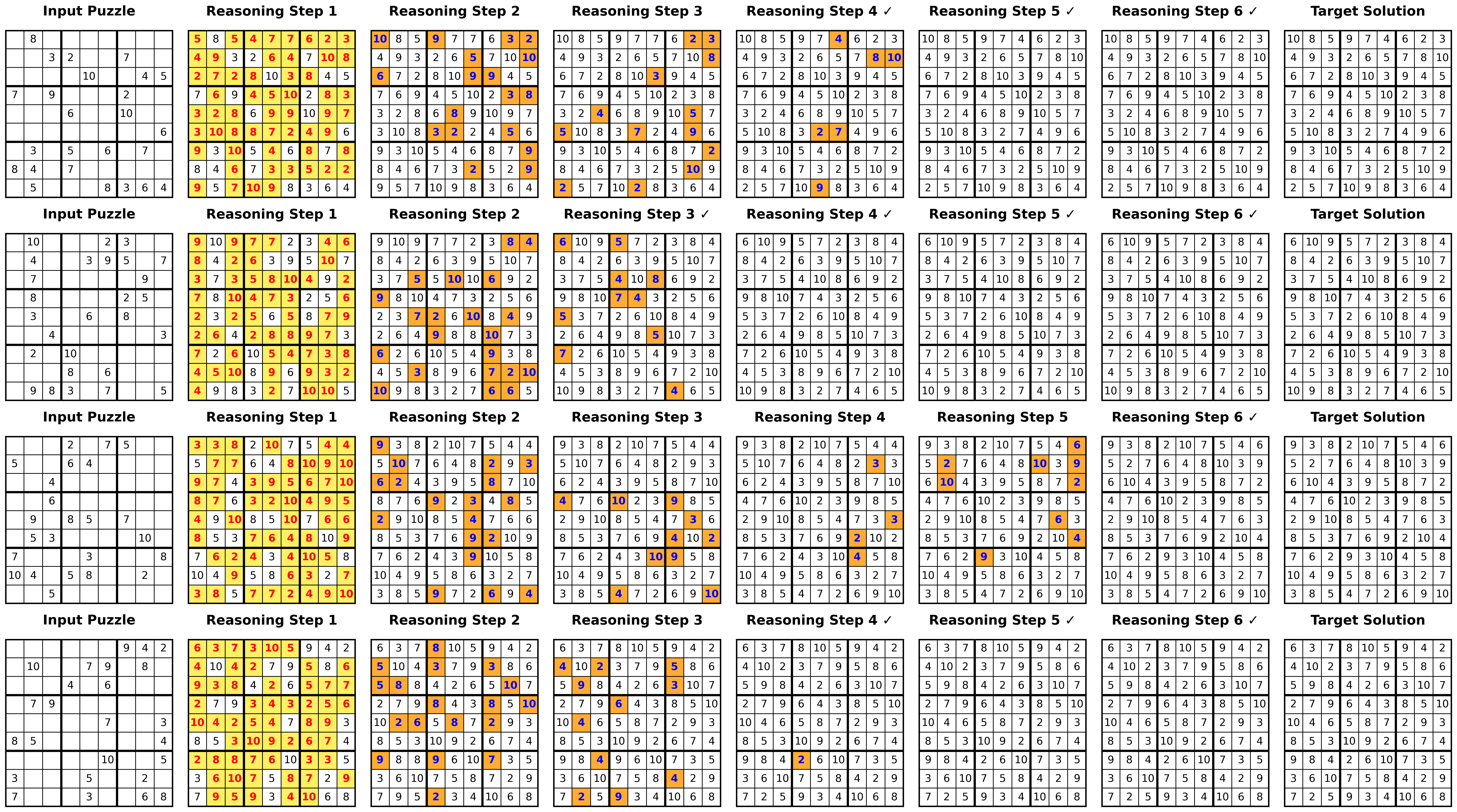}
    \caption{Some concrete examples of diffusion-like reasoning with very few steps. This is in sharp contrast to how humans reason, posing an interesting question on the nature of reasoning.}
    \label{fig:sudoku_grid}
\end{figure}

Figure \ref{fig:sudoku_grid} provides concrete examples where the model halts after only a handful of reasoning steps, while still producing correct solutions. The model dynamically adjusts computation across puzzles and sometimes solves difficult puzzles with very short trajectories. This behavior contrasts with human reasoning, where solving a Sudoku puzzle typically involves many step-by-step constraint propagations until the grid is fully filled. This raises interesting questions about how neural models represent and execute reasoning.

\section{Conclusion}
\label{conclusion}
In conclusion, our investigation reveals several critical insights into the HRM model. First, we found that the H module does not significantly contribute to overall performance; in fact, an HRM model utilizing only the L module performs comparably to the original HRM, consistent to what is reported by \cite{arc_HRM}. Second, the one-step gradient approximation employed by HRM is effectively equivalent to the training paradigm used in diffusion models. Finally, our analysis indicates that ACT does not enhance inference performance; rather, running the model for the maximum number of reasoning steps yields the best results. This raises an intriguing question regarding the recurrent nature of HRM: is it genuinely recurrent, or does it function similarly to a very deep feedforward model network. These findings challenge existing assumptions about HRM and suggest avenues for further research.




\bibliography{bibfile}
\bibliographystyle{iclr2026_conference}


\end{document}